\pdfoutput=1

\documentclass[11pt]{article}

\usepackage{naacl2021}

\usepackage{times}
\usepackage{latexsym}

\usepackage[T1]{fontenc}

\usepackage[utf8]{inputenc}

\usepackage{microtype}

\usepackage{todonotes}
\usepackage{xspace} 
\usepackage[capitalize,noabbrev]{cleveref} 
\usepackage{refcount} 
\usepackage[multiple]{footmisc} 
\usepackage{makecell}
\usepackage[shortcuts]{extdash} 

\title{Machine-Assisted Script Curation}

\author{
  {\bf Manuel R. Ciosici} {\bf Joseph Cummings}, {\bf Mitchell DeHaven}, {\bf Alex Hedges}, {\bf Yash Kankanampati}, \\ {\bf Dong-Ho Lee}, {\bf Ralph Weischedel}, {\bf Marjorie Freedman} \\
  \texttt{manuelc@isi.edu},~~\texttt{weisched@isi.edu},~~\texttt{mrf@isi.edu} \\
  Information Sciences Institute, University of Southern California
}

\newcommand{\gpttwo}{GPT\=/2\xspace}
\newcommand{\thecuratorL}{Machine-Aided Script Curator~(MASC)\xspace}
\newcommand{\thecurator}{MASC\xspace}
\newcommand{\dashspace}{\hspace{1 pt}/\hspace{1 pt}}

\begin{document}
\maketitle
\begin{abstract}
We describe \thecuratorL, a system for human-machine collaborative script authoring. Scripts produced with \thecurator include (1)~English descriptions of sub-events that comprise a larger, complex event; (2)~event types for each of those events; (3)~a record of entities expected to participate in multiple sub-events; and (4)~temporal sequencing between the sub-events. \thecurator automates portions of the script creation process with suggestions for event types, links to Wikidata, and sub-events that may have been forgotten. We illustrate how these automations are useful to the script writer with a few case-study scripts.

\end{abstract}

\section{Introduction}

Scripts have been of interest for encoding procedural knowledge and understanding stories for over 40 years~\cite{schankscripts}. In the form of checklists, recording procedural knowledge has revolutionized ﬁelds like medicine and aviation by encoding expert knowledge and best practices~\cite{Degani_Wiener_1993,gawande2010the}. In the last few years, researchers have turned their attention to automatic script discovery from text~\cite{chambers-2013-event,weber-etal-2020-causal,weber-etal-2018-hierarchical}. However, exclusively data-driven sub-event discovery methods face the challenge that narrative descriptions often omit common knowledge.\footnote{Common knowledge might be missing from narrative descriptions due to the \emph{quantity} and \emph{relevance} maxims~\cite{grice1975logic}.}

We aim for a process for building a library of scripts through human-machine collaboration leveraging NLP techniques to augment human background knowledge. The resulting demonstration system serves two related purposes. First, it is a knowledge acquisition tool that supports the development of a repository of scripts for use by downstream applications. Second, it is an annotation tool that supports the creation of a library to aid our understanding of how people create scripts. Such a library can inform and/or benchmark future script discovery approaches. Each script includes a natural language description of the steps in the complex event with links to an ontology. Events within a script are connected by (a)~temporal order (e.g., negotiating the price of a car happens \emph{before} buying the car) and (b)~by shared argument (e.g., the person buying a car is also the person who negotiated its price). We designed \emph{\thecuratorL}, our script-creation tool, to be used by non-NLP experts.

While approaches to script discovery suffer from the incompleteness of text, human attempts to write machine-interpretable scripts suffer from the writer's own tendency to omit steps and, where required, the challenge of mapping to a formal ontology. To assist the script creators, \thecurator makes three types of suggestions: (1)~the ontological type for each event; (2)~a fine-grained ontological type for suggested arguments; and (3)~steps that the curator might have forgotten.

In the following sections, we describe the process of creating a script in \thecurator and the NLP components that support suggestions.\footnote{A video of \thecurator is available at \url{https://youtu.be/slvZWAYkRmA}, and the source code and the sample scripts are at \url{https://github.com/isi-vista/MASC}.} While a large-scale script repository is beyond this paper's scope, we have created five sample scripts, which we use as case studies for understanding the script creation process and the suggestion capabilities. In \Cref{sec:capabilities}, we use these scripts to measure the utility of \thecurator's suggestion capabilities. In \Cref{sec:discussion}, we describe the scripts' characteristics.

\section{Related Work}

\citet{schankscripts} proposed organizing knowledge about human behavior using scripts. Recent approaches attempt to ``induce'' scripts from large amounts of data rather than write scripts manually~\cite{rudinger-etal-2015-script,weber-etal-2018-hierarchical}. Although improving year over year, these models still perform poorly (Recall@100 of \textasciitilde7\%, \citealp{weber-etal-2020-causal}) at predicting next events, given a set of preceding events - a necessary building block of scripts. These models' training data was obtained by asking human annotators to decide if event B happened because of event A. In contrast, the scripts produced by our curation tool incorporate the complexities of many different events in various causal orderings.

Both symbolic and neural approaches suffer from the lack of generic knowledge to ``fill-in-the-blanks'' or reject impossible events. Training systems to incorporate common-sense knowledge~\cite{lin-etal-2019-kagnet, shwartz-etal-2020-unsupervised} has not yet addressed script creation. Another source of information for script discovery could be extraction from multiple languages and modalities. While some extraction systems have incorporated these other sources~\cite{li-etal-2020-gaia}, such extractions have not yet fed into script discovery. Resolving the co-occurrence of events or entities between languages and modalities often relies on a common mapping, e.g., a structured ontology, such as ACE~\cite{ace2005} or ERE~\cite{song-etal-2015-light}. While our \thecuratorL does employ a structured ontology, it does not currently incorporate multi-modal or non-English sources. However, the limited ontology allows the event-sequencing background knowledge we encode to be used as a supplement to state-of-the-art information extraction systems, like OneIE~\cite{lin-etal-2020-joint} and DYGIE++~\cite{wadden-etal-2019-entity}, providing connections between otherwise disconnected extractions.

\label{sec:overview}
\begin{figure}
    \centering
    \includegraphics[width=0.47\textwidth]{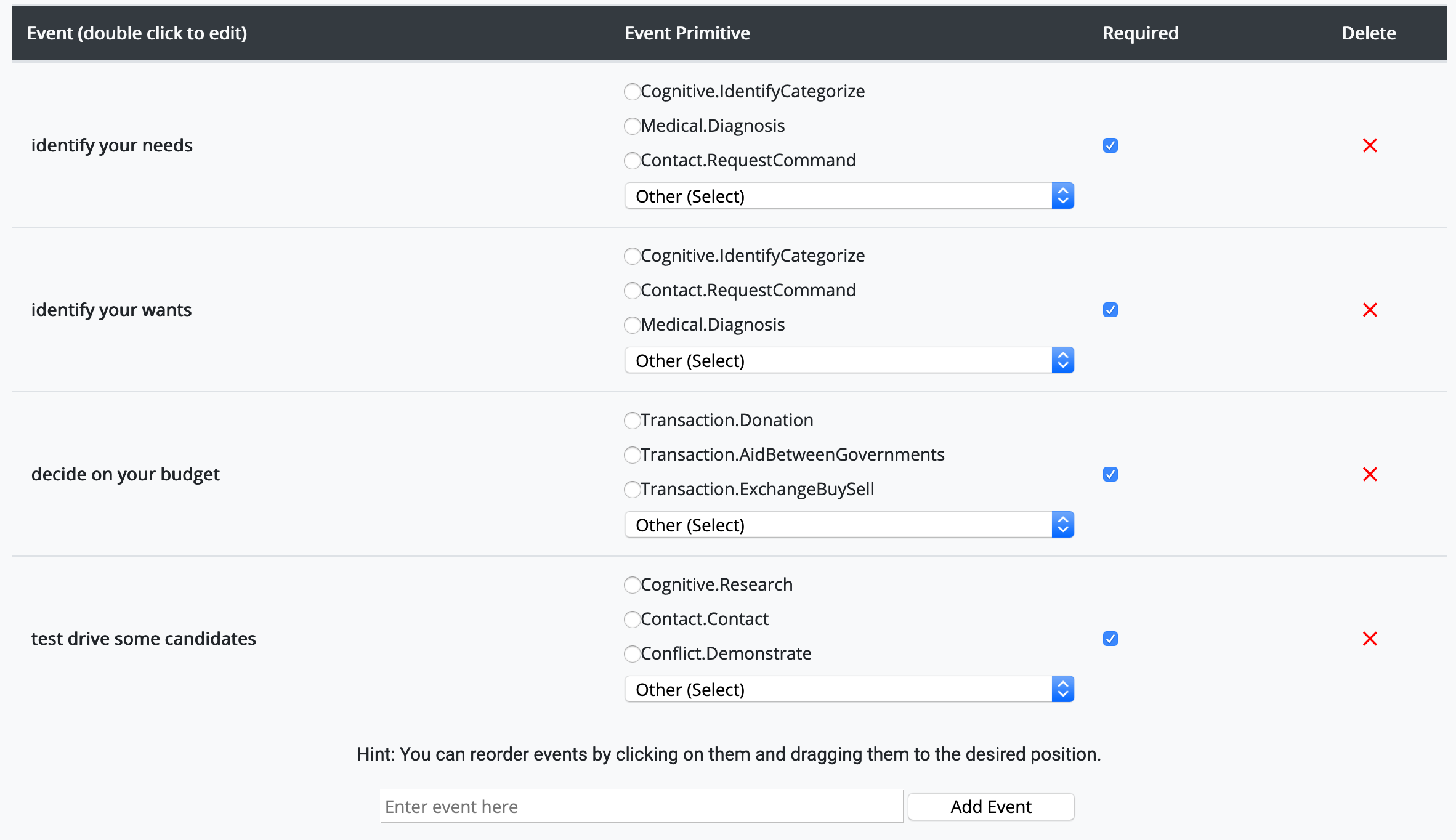}
    \caption{Adding events to the \emph{buying a car} script.}
    \label{fig:adding_events}
\end{figure}

\section{Overview of Script Creation}

The curator initiates script creation by providing a name and description for the script and then enters, as text, the events in the script~(\Cref{fig:adding_events}). Step entry is free-form, but we have noticed a tendency for curators to enter short, imperative sentences around a central agent's actions (e.g., \emph{go to a car dealership}, \emph{take a test drive}). Currently, script creation, unlike traditional annotation, is decoupled from any particular document. In cases where the curator is not familiar with a topic, we have used external resources to provide context (e.g., a Wikihow page open in a different window). In this setting, curation is akin to annotation that encourages the annotator to use both the material they read and prior knowledge.

The curators assign an ontology type to the main event in each step~(e.g., \emph{Movement} for both \emph{go to a car dealership} and \emph{take a test drive}). The ontology is configurable and can be replaced. We include a project-specific ontology with \thecurator's source code. When saved, scripts include both the curators' description and the selected ontology type (described in \Cref{sec:event_types}). This choice allows type decisions to be revisited if the ontology changes and limits the degree to which the small number of event types constrains the script's expressiveness. Downstream applications can choose whether to use the linguistic representation of the events or the normalized ontology types.

\begin{figure*}
    \centering
    \includegraphics[width=0.95\textwidth]{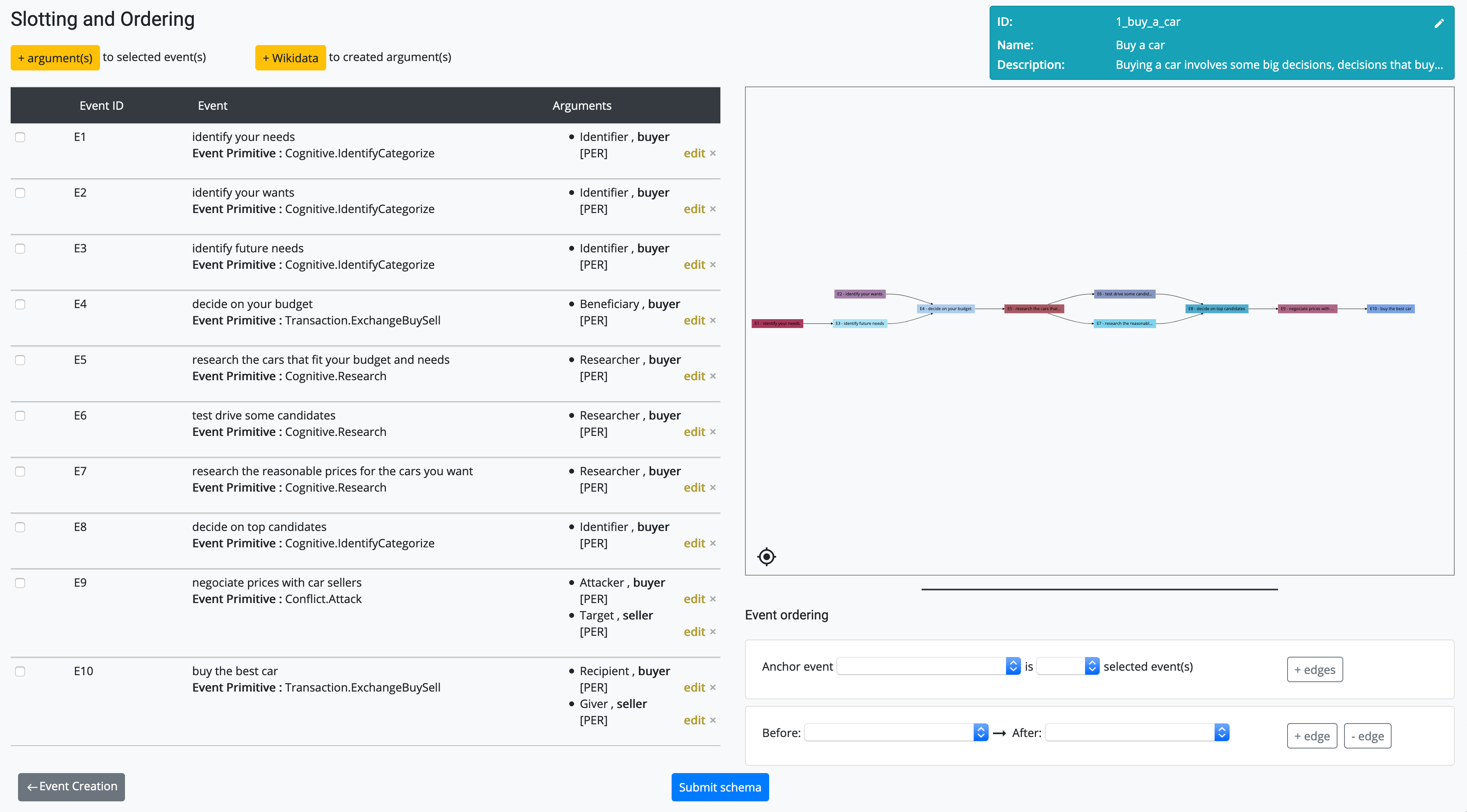}
    \caption{Adding details to events. For each event on the left, curators can add arguments. On the right side, curators can establish temporal order and visualize the script as an interactive graph.}
    \label{fig:slotting_and_ordering}
    \vspace{-0.3cm}
\end{figure*}

After the curators finish entering events, they encode connections between the events~(\Cref{fig:slotting_and_ordering}). There are two ways to connect events: the ﬁrst, traditionally the focus of scripts, is temporal order; and the second is shared arguments (e.g., the same person is the agent of both \emph{Movement} events \emph{go to a car dealership} and \emph{take a test drive}). To add sequential order, the curators enter pairwise \emph{before} relations. Alternatively, they select multiple events and anchor them as coming before or after a single event. The latter method is convenient when the complete order is under-defined.\footnote{For example, after arriving at the car dealership, the potential buyer is likely to both walk around looking at cars and talk to a salesperson, but there is no defined order between the walking and talking.} The curators add shared arguments to the script by selecting multiple events with the same argument, naming the argument (e.g., \emph{buyer, seller} in \Cref{fig:slotting_and_ordering}), and assigning an entity type (e.g., \emph{PER} in \Cref{fig:slotting_and_ordering}) and ontological role to each argument (e.g., \emph{Identifier, Researcher} in \Cref{fig:slotting_and_ordering}). While this process is mostly manual, \thecurator uses the ontology's constraints to limit the available label options. In addition to project-specific entity types, \thecurator suggests links to the much larger set of types available using Wikidata entities (e.g., suggesting \href{https://www.wikidata.org/wiki/Q786803}{Q786803} for \emph{car dealership}). These links provide a connection to an extensive knowledge graph and can provide additional information when the scripts are applied.

Finally, the curators review events that are automatically generated based on the manually entered description and initial script (described in \Cref{sec:event_recommendations}). The suggestions can add intermediate steps that the curators may have missed, complete a script that was intentionally unfinished by the curator, or suggest alternative related paths (e.g., leasing instead of purchasing a car).

\section{Suggestion Capabilities}
\label{sec:capabilities}
To aid script creation, \thecurator incorporates three suggestion capabilities: suggestions for the ontological event type, suggestions for links to Wikidata, and suggestions for additional events to incorporate in the script. Below, we describe the models behind these capabilities and, for each model, report the accuracy using the five sample scripts created for this paper. Given the small sample size, the five sample scripts are best thought of as case studies, not a benchmark. \Cref{tbl:five_scripts} provides per-script analysis.

\subsection{Event Type Classification}
\label{sec:event_types}

Each sub-event is ontologized with one of 41 event types through a semi-automated process. The ontology labels support connecting information to extraction engines and thus allow a script to provide potential event-event relations given information extraction output. Furthermore, the ontology labels provide language- and media-independent knowledge for identifying potential instances of the scripts.

There has been much work on automatic detection of event types (and triggers) in text (e.g., \citet{bronstein-etal-2015-seed, lin-etal-2020-joint, peng-etal-2016-event}). Here, our input data (and goals) are slightly different. The ontology we use, while overlapping with ACE~\cite{ace2005}, introduces several new event types for which we do not have annotated training examples. Instead, the ontology provides a short definition and template for each event type. The curator's input events tend to be short imperative sentences with different linguistic characteristics than the text annotated in, e.g., ACE. Unlike standard information extraction, we need not identify a specific trigger phase.\footnote{Triggers are often used as a means to identify arguments of interest. But here, partly because of the telegraphic nature of the text entries, the arguments are often missing and, therefore, explicitly added.} Thus, we use a different approach to event labeling.

To map from the curators' description of an event to the ontology, we use a version of Sentence-RoBERTa~\cite{reimers-gurevych-2019-sentence}\footnote{Our Sentence-RoBERTa model is trained on more data. We use the two data sets in the original paper, SNLI~\cite{bowman-etal-2015-large} and MNLI~\cite{williams-etal-2018-broad}, and add the newer ANLI~\cite{nie-etal-2020-adversarial}.} to estimate the similarity of the curators' text input to the prose description of each action in the ontology. For example, for the user input \emph{go to a car dealership}, the action description \emph{Explicit mention of granting or allowing entry or exit from a location} receives the highest similarity score, and the corresponding action type \emph{Movement.Transportation} becomes one of the recommendations. \thecurator suggests the three ontology actions most similar to the user's description. The user can accept one of the suggestions or pick a different type from the ontology~(\Cref{fig:adding_events}, second column).

As mentioned earlier, the event type similarity depends on the ontology event type definitions and the event type templates. In preliminary experiments, we found using both together outperformed using either only the definitions or only the templates. While \thecurator's event type classification does not require training data, it depends on both the presence of templates and definitions in the ontology and their quality.

\textbf{Performance on Case-Study Scripts} The five scripts contain 58 events. We measure how often the model correctly predicts the event type that the curator selects. Accuracy of the top-1, -3, and -5 are 24, 48, and 55, respectively.\footnote{The mean reciprocal rank (MRR, \citealp{radev-etal-2002-evaluating}) was $0.35$ on the top three model predictions.} \thecurator presents the top-three suggestions to the curator; thus, accuracy at top-3 most closely relates to the curator's experience.

\subsection{KGTK}
\label{sec:entity_labeling}
\begin{figure}
    \centering
    \includegraphics[width=0.47\textwidth]{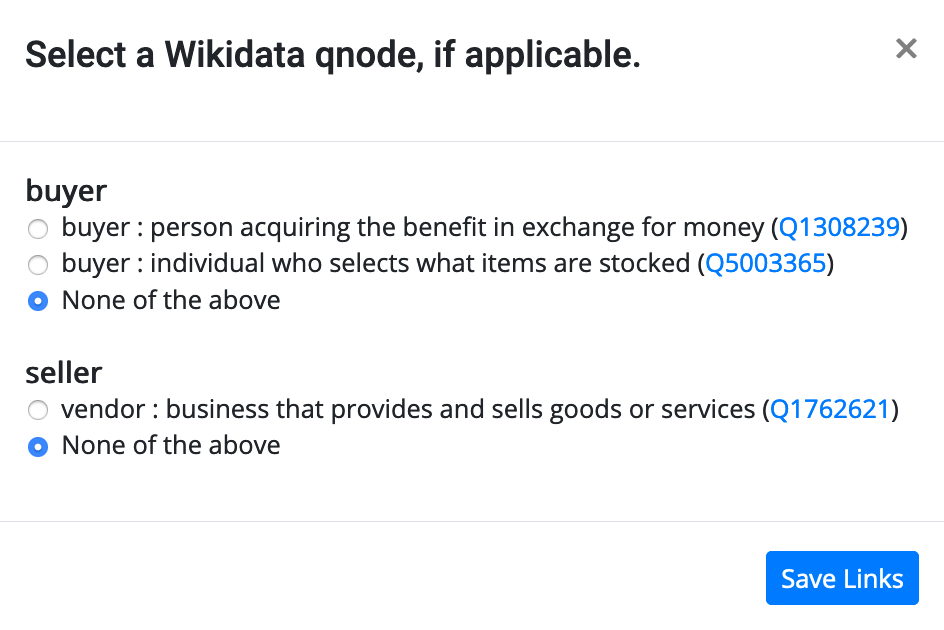}
    \caption{Reviewing the Wikidata link suggestions.}
    \label{fig:entity_linking}
\end{figure}

In \Cref{sec:overview}, we describe identifying the key repeating arguments of script events and labeling those arguments with their entity type and their role in each event using an ontology. That ontology provides only coarse distinctions between entities (e.g., a single category for facilities that does not distinguish \emph{a car dealership} from \emph{a school} or \emph{a bank}). To support finer-grained distinctions and, in the future, leverage external knowledge sources, we incorporate connections to Wikidata\footnote{\url{https://www.wikidata.org}} using KGTK~\cite{kgtk}. \thecurator's links aim to ground descriptive noun phrases~(e.g., \emph{car dealership}) in the large Wikidata ontology and do not require grounding specific, named entities~(e.g., \emph{Toyota}).

KGTK is an open-source toolkit that simplifies searching and interacting with various knowledge graphs, including Wikidata. KGTK provides a simple API for searching Wikidata entries, via Elasticsearch,\footnote{\url{https://www.elastic.co/elasticsearch/}} based on their titles and aliases~(e.g., the Wikidata entry \emph{motor car} also has the aliases \emph{auto}, \emph{automobile}, and \emph{car}). KGTK also provides filtering functionality for candidate Wikidata entries. Since we are not interested in grounding specific named entities, we only return Wikidata entries representing Wikidata classes. Within \thecurator, KGTK allows users to link terms used in events to Wikidata. During argument creation, the curator provides a text label for each key argument. A background process then queries KGTK using the text label assigned to each argument. Candidates from KGTK are reranked using the Sentence-RoBERTa model to generate similarity scores between the label strings and the candidate Wikidata text descriptions. Before finishing a script, for each term in the script, the curator can select one of the candidates from KGTK or \emph{None of the above}~(\Cref{fig:entity_linking}).

\textbf{Performance on Case-Study Scripts}. To evaluate entity linking, we treat the scripts created by the curators (and the mapping from the reference variables to Wikidata) as the labels. This is necessary since we do not have a ground-truth mapping from strings to Wikidata entities, and curators can use the same string to reference different entities. For example, \emph{car} can refer to an automobile, a railway carriage, or a streetcar. The metric we use measures the ratio of reference variables linked to a speciﬁc Wikidata entity to the total number of reference variables used. We ﬁnd that curators link 67\% of the unique reference variables to Wikidata (e.g., \emph{buyer} in \Cref{fig:entity_linking}). We have not measured the ceiling on using Wikidata as an argument ontology. However, we suspect that refining the linking approach could yield more connections to Wikidata. Even at this low level of recall, at least a few concept-speciﬁc elements match for most scripts. In the future, these connection points could support script augmentation using common-sense and domain knowledge from Wikidata.

\subsection{Event Recommendations}
\label{sec:event_recommendations}
\begin{figure}
    \centering
    \includegraphics[width=0.47\textwidth]{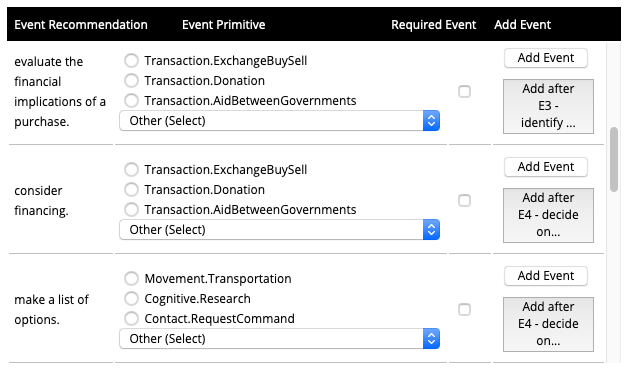}
    \caption{\gpttwo recommendations for \emph{buying a car}.}
    \label{fig:gpt2_recommendations}
\end{figure}

Since even the most experienced curators may overlook an action in an event script, we explored hypothesizing omitted events using \gpttwo~\cite{radford2019language} \textit{without any fine-tuning}.

The first challenge is formulating input to \gpttwo. We provide the title/name of the schema (e.g., \textit{buying a car}), a description of the complex event (e.g., \textit{Purchasing a car is a large investment that requires careful documentation and consideration of transportation requirements.}), and a request (e.g., \emph{Describe steps of buying a car.}), followed by the first few events of the script. In the initial version, we used a form of the events as \textit{First, Identify your needs. Then, Decide on your budget. Next, Identify car models you can afford.} However, a numerical formulation proved much more effective (e.g., \textit{1.~Identify your needs 2.~Decide on your budget 3.~Identify car models you can afford 4.}) and resulted in more coherent events.

To filter undesirable or redundant output, we pass \gpttwo outputs through a sequence of filters. We remove undesired strings characteristic of neural text generation, like empty strings~\cite{stahlberg-byrne-2019-nmt}, and outputs that are invalid in the context of schema creation: strings of less than two words and those with sequences of non-alphabetic characters. We address duplicated output, a considerable concern for \gpttwo, especially given the short and similar inputs.\footnote{\gpttwo often generates strings with a similar meaning, but lexically different, e.g., for a script on buying a car, it might generate \emph{buy}, \emph{buy the car}, and \emph{purchase the car}. It is superfluous to show users all three suggestions.} The filters eliminate strings with duplicates in the alternatives or the human-curated schema. To account for semantic duplicates, such as \textit{go to dealership} and \textit{go to the car dealership}, we use a variant of Gestalt Pattern Matching~\cite{ratcliff1988pattern} through Python's \textit{difflib}. For usability, we suggest at most 12 sub-events per script. \Cref{fig:gpt2_recommendations} shows the interface for reviewing event recommendations.

\textbf{Performance on Case Studies.} We measure the performance of \gpttwo recommendations in two ways. First, we generate recommendations for ﬁve scripts created by curators and ask the curators to accept relevant \gpttwo recommendations. We instruct curators to accept recommendations even if the recommended events represent alternative paths (or are semantically redundant). With these instructions, the curators accept $98\%$ of \gpttwo's recommendations. The high acceptance rate indicates that even with our simple setup for event recommendation using a language model, the system suggests domain-relevant events.

For the second evaluation, we instruct the curators to accept only those \gpttwo recommendations that add to their existing script. In other words, they only accept events that add details to the scripts or supply some missing information. We instruct curators to reject recommendations for alternative script scenarios. With these instructions, curators accept $23\%$ of \gpttwo's recommendations. This result illustrates the feasibility of supplementing human knowledge with generations from language models. Since \thecurator uses \gpttwo \emph{after} the human felt the script was complete, the machine identifies events previously overlooked by the human.

\textbf{Mixed-Initiative script curation.} Given the success of \gpttwo recommendations after script curation, a natural next step is for curators to work with \gpttwo interactively. In the \emph{mixed-initiative mode}, a curator specifies a script's name, definition, and first step. \gpttwo then suggests multiple options for the next step. The curator can use one of the suggestions, edit it, or ignore all the suggestions and manually input the next step. Every time the curator adds a step to the script, \gpttwo follows with suggestions for the next step. We found that automated step generation took less than 3 seconds in the slowest case on modern hardware (NVIDIA GeForce RTX2080Ti).

\begin{figure}
    \centering
    \includegraphics[width=0.47\textwidth]{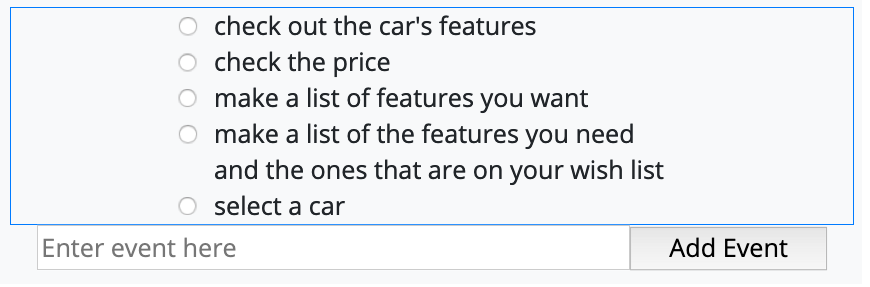}
    \caption{Mixed-Initiative: \gpttwo's suggestions for the script \emph{buying a car}, given the first step \emph{Identify your needs}.}
    \label{fig:gpt2_mixed_initiative}
    \vspace{-0.3cm}
\end{figure}

To evaluate the effectiveness of mixed-initiative mode, we asked four curators to create a total of twelve scripts using the mode. We instructed the curators to accept event suggestions only when they are a natural continuation of the script. Out of \gpttwo's 105 suggestion sets, the curators accepted an event from 50 sets ($48\%$ acceptance rate). In six more cases, the curators used a \gpttwo suggestion as a starting point and edited the suggestions to suit the script better. We found the mixed-initiative scripts to be just as comprehensive as the scripts detailed in \Cref{tbl:five_scripts}, where \gpttwo suggested missing events only after the curators created an initial script.

\section{Discussion and Future Work}
\label{sec:discussion}

\begin{table*}
    \centering
    \resizebox{\textwidth}{!}{
    \begin{tabular}{rl|c|c|c|c|c}
        & & EVAC & FOOD & JOB & MED & MERGER\\ \hline
       1 & \# Events in initial script & 16 & 9 & 16 & 5 & 12 \\ \hline
       2 & Accuracy at top-1, 3, 5 for event types & 25\dashspace44\dashspace50 & 11\dashspace33\dashspace67 & 13\dashspace44\dashspace44 & 20\dashspace60\dashspace60 & 50\dashspace67\dashspace67 \\ \hline
        3 & \makecell[l]{\# Entity instances, occurrences of those \\ entities, and unique links to Wikidata} & 2\dashspace26\dashspace1 & 5\dashspace18\dashspace3 & 2\dashspace24\dashspace2 & 3\dashspace11\dashspace3 & 3\dashspace24\dashspace2 \\ \hline
        4 & \makecell[l]{\# Event suggestions selected for single \\ script and all relevant (max. 12 per script)} & 4\dashspace8 & 3\dashspace12 & 0\dashspace11 & 5\dashspace12 & 2\dashspace12 \\ \hline
        5 & Non-linear path & Y & Y & Y & N & N \\ \hline
        6 & Self-reported time & 1.5 hrs & 0.5 hr & 1 hr & 0.5 hr & 2.5 hrs \\ \hline
    \end{tabular}
    }
    \caption{Characteristics of five sample scripts.}
    \label{tbl:five_scripts}
\end{table*}

With this demonstration system, we provide an approach to human-machine collaboration for building a repository of scripts. Having such a repository, for a diverse set of events, will allow us to investigate how procedural knowledge introduced to the AI community 40 years ago~\cite{schankscripts} can be broadly applied. By facilitating the human creation of scripts, we can better understand what is required to develop automatic script discovery approaches.

While we have not yet created a large repository of scripts, we have created ﬁve scripts with which we start this analysis. The scripts cover topics with varying degree of ``common knowledge'': \emph{Planning and Managing an Evacuation} (EVAC), \emph{Ordering Food at a Restaurant} (FOOD), \emph{Finding and Starting a New Job} (JOB), \emph{Obtaining Medical Treatment} (MED), and \emph{Corporate Merger or Acquisition} (MERGER). A single curator created these scripts, which we use to illustrate future directions for \thecurator and interesting properties of the scripts themselves. Having multiple curators for even a small number of scripts would provide insights into the diversity, prior knowledge, and level of detail a script author uses. In our analysis, we have seen that the scripts created with \thecurator encode knowledge that is uncommon in news-like data sets. For example, our curator included \emph{sign confidentiality agreement} as an event in the script for a MERGER. While news frequently reports the final step of a merger, the full process is rarely described.

\Cref{tbl:five_scripts} summarizes the key characteristics of each of these scripts. They vary in (a)~the number of steps initially created (row 1), with only 5 steps for MED and 16 for both EVAC and JOB; and (b)~the time required for initial script creation (row 6). The script that took the longest was not the one with the most steps (or the most arguments). Instead, it was the domain that the curator knew the least about (and thus chose to research). For all five scripts, there were cases where the event type suggestions were correct, but for three of the five, \thecurator suggested the correct type less than half the time, suggesting that better automatic event typing could increase the curators' speed.

All scripts contain entities that play a role in multiple events (row 3, first and second numbers). For example, in EVAC, the evacuation manager plays some role in all events, while the evacuee plays a role in most but not all. While some arguments cannot be linked to Wikidata, all five scripts contain at least one argument that can be linked (row 3, last number). Future work could both improve linking accuracy and use Wikidata as a source of knowledge to provide additional context (and suggestions) to the curator.

While the prototypical script is a timeline with complete order between all pairs of events, we see sub-graphs with unordered steps in our data. Three of the five sample scripts display this behavior; for example, in JOB, \emph{searching for open positions} and \emph{notifying network that they are looking for a job} are unordered. The visualization of the schema in \Cref{fig:slotting_and_ordering} illustrates this pattern with no order between \emph{E2} and \emph{E3}.

\thecurator incorporates machine suggestions of unrecorded events. In four of the five scripts, the curator accepted at least one suggestion. Interestingly, the curator incorporated more suggestions for two events that one thinks of as everyday experiences (FOOD and MED) than they did for the script they were unfamiliar with (MERGER). This suggests that the recommendation functionality can be useful even in a familiar domain; by capturing what the curator omits through forgetfulness or because they assume common knowledge. Further exploration of how a machine can aid a person whose knowledge may be incomplete or may forget to be explicit seems promising. Examples of possible research directions include incorporating suggestions from approaches that discover scripts (e.g., \citet{rudinger-etal-2015-script,weber-etal-2018-hierarchical,weber-etal-2020-causal}) and leveraging background knowledge (e.g., Wikidata).

\section{Ethical Considerations}
 
Many technological innovations require ethical considerations, even more so for those involving machine learning while also being a demonstration paper that provides working technology. Below we address the review questions raised in the NAACL Ethics Review Questions.\footnote{\url{https://2021.naacl.org/ethics/review-questions/}}

\emph{Bias.} The bias in generative language models has been well documented. In general, using a human-in-the-loop process means that rather than treating an automatically generated label or event as correct, we treat it as a suggestion that the curator can ignore. Still, the suggestions can inﬂuence the curator. Thus it is vital that the metrics reported in this paper be interpreted with an understanding of the potential for bias and any use of \thecurator account for bias.

\thecurator incorporates both a predefined ontology and the ability to link to an extensive external resource (Wikidata). Given the size of the predefined ontology is small, to apply \thecurator to a new domain, users would likely need to update the ontology. \thecurator's approach to aligning English descriptions to the ontology makes adding new event classes easy. Wikidata, while much larger and growing, is also subject to the bias of Wikidata's editors, their knowledge, and their choices about what to include. Wikidata over-represents some issues, while some socially important ones are under-represented or missing. Wikidata linking is optional; thus in a domain that is not well covered, a curator can skip the linking step or replace Wikidata with a domain-relevant resource.

The suggestion capabilities described in \Cref{sec:capabilities} use pretrained language models (\gpttwo and RoBERTa). The bias of these algorithms, measuring that bias, and mitigating it is an active area of work. Recent work has provided data sets for measuring bias \cite{Nadeem2020} and meta-studies of the approaches taken to study and address bias \cite{blodgett-etal-2020-language}. Much work has focused on bias as it impacts demographic groups. \thecurator focuses on events, not individuals. The publicly available \gpttwo models have learned from data that might not cover current events (e.g., \gpttwo was trained before the COVID-19 epidemic), represents only English dialects from the inner-circle~\cite{dunn-adams:2020:LREC}, and contains toxic language~\cite{gehman-etal-2020-realtoxicityprompts}. In our immediate context, we mitigate against the challenge presented by language model bias by requiring manual review of all automatically suggested output. If the ideas in this paper were extended to a fully automatic approach, language model and domain-specific studies of the impact of bias on LM-based suggestions would be necessary.

\emph{Data Set.} To understand how the tool is used and future research directions, we created five sample scripts which we included in the supplementary material. These scripts provide interesting examples of what we could learn from a larger scale data set; however, they are not large enough themselves to serve as a new benchmark. The five scripts were created by full-time research staff compensated following US state and federal law. The scripts were created by a single individual and represent that individual's pre-existing knowledge (and their implicit biases). To counter bias in a large-scale script repository, we recommend that the curator workforce is diverse and that any given activity is represented in scripts written by multiple people. Any released repository should have sufficient reporting about the data set creators to provide users with an understanding of data bias. The paper reports empirical results based on this five script sample. However, the paper acknowledges that the sample is small and treats these results as case studies for \thecurator, not a new benchmark.

\emph{Intended Use.} The most immediate use of \thecurator is to create a repository of script information -- either broadly available to researchers or within a specific research community. In some cases, e.g., the steps to plan a rescue operation, both the generation of the script and its application are generally understood as positive. In other cases, e.g., the steps in grooming an individual for human trafficking, the script's conclusion is negative, but understanding the process is necessary to prevent the activity. As AI's ability to discover and apply such knowledge increases, it will be necessary to regularly audit the use cases to ensure the focus remains a benefit to society. If the human-in-the-loop approaches used here were integrated into a fully automated system, further auditing of bias (and accuracy) would be necessary.

\emph{Compute Time and Power}. Most of the models used for this demonstration are pretrained and publicly available. The pretraining and fine tuning described in \cref{sec:event_types} took less than 20 hours using a single GPU.

\section*{Acknowledgment}
\label{sec:acknowledgment}
This material is based on research supported by DARPA under agreement number FA8750-19-2-0500. The U.S. Government is authorized to reproduce and distribute reprints for Governmental purposes notwithstanding any copyright notation thereon. The views and conclusions contained herein are those of the authors and should not be interpreted as necessarily representing the official policies or endorsements, either expressed or implied, of DARPA or the U.S. Government.

\bibliography{custom}
\bibliographystyle{acl_natbib}

\end{document}